\documentclass[a4paper,fleqn]{cas-dc}

\usepackage[authoryear]{natbib}

\def\tsc#1{\csdef{#1}{\textsc{\lowercase{#1}}\xspace}}
\tsc{WGM}
\tsc{QE}

\newcommand{\norm}[1]{\left\lVert#1\right\rVert}
\DeclareMathOperator*{\argmin}{argmin}

\begin{document}
\let\WriteBookmarks\relax
\let\printorcid\relax
\def\floatpagepagefraction{1}
\def\textpagefraction{.001}

\shorttitle{}    

\shortauthors{}  

\title [mode = title]{NeST: Neighborhood-aware semantic alignment and temporal modulation for LLM based time series forecasting} 

\author[1]{Jayanie Bogahawatte}[
    orcid=https://orcid.org/0009-0001-8788-793X
]
\cormark[1]



\ead{jbogahawatte@student.unimelb.edu.au}

\credit{Conceptualization, Investigation, Methodology, Software, Writing – original draft}

\affiliation[1]{organization={AI, Optimization and Pattern Recognition Research Group},
            addressline={Faculty of Engineering and Information Technology, University of Melbourne}, 
            country={Australia}}

\author[1]{Sachith Seneviratne}[
    orcid=https://orcid.org/0000-0001-9094-2736
]


\ead{sachith.seneviratne@unimelb.edu.au}

\credit{Conceptualization, Writing – review and editing, Supervision}

\author[1]{Maneesha Perera}[
    orcid=https://orcid.org/0000-0002-9779-0699
]


\ead{maneesha.perera1@unimelb.edu.au}

\credit{Conceptualization, Writing – review and editing, Supervision}

\author[1]{Saman Halgamuge}[
    orcid=https://orcid.org/0000-0002-2536-4930
]
\ead{saman@unimelb.edu.au}

\credit{Conceptualization, Writing – review and editing, Supervision}

\cortext[1]{Corresponding author: Jayanie Bogahawatte (jbogahawatte@student.unimelb.edu.au), Postal address: Department of Mechanical Engineering, The University of Melbourne, Victoria 3010, Australia}

\begin{abstract}
Adapting Large Language Models (LLMs) trained on discrete text data, to forecast continuous time series signals is challenging. While finetuning the LLMs enables such adaptation, effectively integrating both textual and time series information in the prompt is critical. Current LLM-based time series forecasting methods combine the two modalities through simple concatenation or parameter heavy cross-attention. Moreover, existing methods embed time series data using decomposition techniques that may inadequately capture complex temporal dynamics. To address these limitations, we propose neighborhood-aware semantic alignment and temporal modulation based framework (NEST) to formulate a new text-integrated time series prompt to finetune the LLM. First, we generate neighborhood-aware text prototypes that are optimized to represent local neighborhoods of pretrained word token embeddings of the LLM. Second, we align them with temporal representations of the time series input using a nearest-neighbor contrastive objective, after which the top-k most relevant text prototypes are retrieved. Third, we introduce text prototype conditioned temporal modulation that uses the retrieved text prototypes to adaptively scale and shift time series features. Extensive experiments demonstrate that NeST consistently outperforms state-of-the-art methods across eight benchmarks, achieving an average 1.2\% reduction in MSE for long-term forecasting. In addition, it demonstrates strong generalization, yielding an average 4.9\% reduction in MSE in zero-shot forecasting. Beyond benchmark datasets, NeST also delivers robust performance on a real-world distributed photovoltaic power forecasting task across nine datasets, improving the average R$^2$ score by 3.3\%. These findings demonstrate the effectiveness and generalizability of NeST for adapting LLMs to diverse time series forecasting tasks.
\end{abstract}



\maketitle

\section{Introduction}
\label{sec:Introduction}

Time series forecasting is a pivotal analytical task with applications spanning across retail, finance, healthcare, meteorology, energy, and transportation, where accurate predictions are crucial for decision making and strategic planning \cite{stirling2021seizure,hong2020energy,perera2024day}. Classical time series forecasting techniques including autoregressive integrated moving average (ARIMA), exponential smoothing, and Theta \cite{brockwell2002introduction,makridakis2018m4} have been frequently used for such predictive analytics. With the recent advances in deep learning, particularly Recurrent Neural Networks (RNNs), Long Short-Term Memory networks (LSTMs) and more recently transformer-based frameworks have shown significant promise in time series forecasting \cite{benidis2022deep}. However, due to the domain-variant characteristics, a model designed for a specific time series dataset is not typically generalizable for other time series datasets \cite{jiang2024empowering}. 

Large foundation models have captured significant interest in several fields such as natural language processing ~\cite{touvron2023llama,radford2019language,raffel2020exploring} and computer vision \cite{kirillov2023segment,caron2021emerging}. However, developing a general-purpose foundation model for time series analysis presents unique challenges. Time series data comes from diverse domains including healthcare, finance, transportation, and energy, which can cause the data to be domain-variant. Additionally, these data can be inherently non-stationary, with underlying statistical properties vary over time. This variability necessitates frequent retraining of foundation models to maintain their effectiveness and this process can be computationally expensive \cite{jiang2024empowering}. Furthermore, the scarcity of large, high-quality publicly available datasets to train foundation models in the time series domain poses a significant problem \cite{zhou2023one}. Due to these challenges, recent work focuses on leveraging Large Language Models (LLMs) for time series forecasting, exploiting the transfer learning capabilities of LLMs and the sequential nature of both text and time series data \cite{jiang2024empowering,jin2024position}. 

LLMs are pretrained on abundant text data, and therefore are well-known to have a robust capability to recognize patterns in sequences \cite{mirchandani2023large}. Even though time series are numerical sequences, adapting LLMs for time series forecasting introduces two significant challenges. First, since LLMs are trained on data from the modality of text, adapting the learned prior of the LLMs for the new modality of time series is required. Second, time series data is continuous in nature in contrast to discrete text data \cite{jin2023time}. Recent work has proposed distinct approaches to address these challenges including unique tokenization methods to generate time series tokens \cite{gruver2024large}, template-based techniques to create sentences from time series data for sentence-to-sentence processing \cite{xue2023promptcast}, reprogramming methods to align time series data with text \cite{jin2023time,sun2023test}, and finetuning LLMs using carefully formulated prompts as input \cite{cao2023tempo,zhou2023one,pan2024s}. Latest approaches demonstrate the importance of effective prompting which integrates both time series and text data when finetuning the LLMs \cite{cao2023tempo,pan2024s}. This enables the LLMs to leverage time series specific information. During the prompt formulation, the large set of word token embeddings of the LLM are utilized to generate a small set of text prototype embeddings. Current methods use different techniques to obtain these text prototypes including linear probing \cite{jin2023time,pan2024s} or fixed, pre-defined prototype sets \cite{sun2023test}. However, these approaches can under-utilize the geometric structure of pretrained word token embeddings, which can cause for degrading performances in few-shot regimes, where text prototypes need to be optimized with limited supervision. In addition, prior approaches often rely on time series decomposition techniques (e.g. additive seasonal-trend decomposition) to learn time series embeddings \cite{pan2024s,cao2023tempo}, yet such decomposition may not be ideal for time series with more complex non-stationary patterns, which are more common in real-world applications \cite{wen2019robuststl}. Furthermore, existing frameworks typically incorporate text prototype conditioned semantic information through token concatenation \cite{pan2024s, cao2023tempo, s2tsllm}, prompt prefixing \cite{zhou2025balm}, or cross attention \cite{jin2023time}. While effective for injecting auxiliary semantics, these approaches directly modify the token sequence, resulting in a rigid coupling between semantic information and temporal representations. Such tight coupling can interfere with the intrinsic temporal structure encoded in time series embeddings and offers limited control over how semantic priors interact with time series features. Additionally, attention based fusion introduces substantial amount of trainable parameters and computational complexity, increasing susceptibility to overfitting, particularly in few-shot learning settings.

To address these limitations, we propose \textbf{Ne}ighborhood-aware \textbf{S}emantic Alignment and \textbf{T}emporal Modulation for LLM based Time Series Forecasting (\textbf{NeST}), a framework that jointly tackles challenges in text prototype learning and integration of both time series and text information in the prompt formulation. First, we introduce a neighborhood-aware prototype learning mechanism that initializes learnable prototypes within the pretrained word token embedding space of the LLM and optimizes them to represent local neighborhoods of these word token embeddings. Second, we introduce a nearest-neighbor contrastive learning (NNCL) objective to align those text prototypes with temporal representations of the time series input. In NNCL, a support queue is maintained as the representative of the full distribution of the text prototypes. This mechanism can yield robust text prototype representations even under limited supervision. Finally, we introduce prototype conditioned temporal modulation, a feature‑wise conditioning mechanism that injects semantic information into temporal representations without altering the token sequence or introducing additional semantic tokens. We divide the time series input into patches of local time windows and the proposed modulation leverages learned text prototypes to generate feature level modulation parameters that adaptively scale and shift time series patch embeddings through residual interpolation. This design preserves the intrinsic temporal structure of time series patch embeddings while enabling smooth and controllable semantic influence. To maintain the pretrained prior of LLMs and reduce computational overhead, we finetune only positional embeddings and layer normalization parameters, keeping the remaining backbone of the LLM frozen. Extensive experiments on a diverse set of benchmark datasets demonstrate that NeST consistently achieves superior performance across both long-term and short-term forecasting tasks. Notably, NeST outperforms existing state-of-the-art methods in few-shot and zero-shot forecasting, highlighting its robustness under data-scarce conditions and strong generalization capability. Beyond standard benchmarks, we further evaluate our method in a real-world energy forecasting setting; distributed photovoltaic (PV) power forecasting, a critical application for grid stability and renewable energy integration. Across nine PV datasets spanning across diverse operating conditions, NeST achieves the best overall performance, substantially surpassing existing approaches. Together, these results establish our method as a general and practical forecasting framework, capable of reliable deployment in both benchmark settings and real-world renewable energy forecasting applications. The main contributions of this work are summarized as follows:
\begin{enumerate}
    \item \textit{Neighborhood-aware text prototype learning}: We introduce a prototype learning mechanism that exploits local neighborhood structure within pretrained word-token embedding space of LLM to learn text prototypes.
    \item \textit{Text and temporal alignment through NNCL}: We introduce an NNCL based objective to optimize the learned text prototypes to align with time series representations.
    \item \textit{Prototype conditioned temporal modulation}: We introduce a feature-wise semantic conditioning mechanism for prompt formulation, that injects textual information into time series representations.
    \item \textit{Comprehensive empirical evaluation}: NeST achieves state-of-the-art performance across long-term, short-term, few-shot, and zero-shot forecasting benchmarks. 
    \item \textit{Evaluation under real-world energy forecasting setting:} We demonstrate the effectiveness of the proposed method on distributed photovoltaic power forecasting, achieving superior performance across multiple datasets originated from distinct geographical locations.
\end{enumerate}

\begin{figure*}
  \centering
  \includegraphics[width=\linewidth]{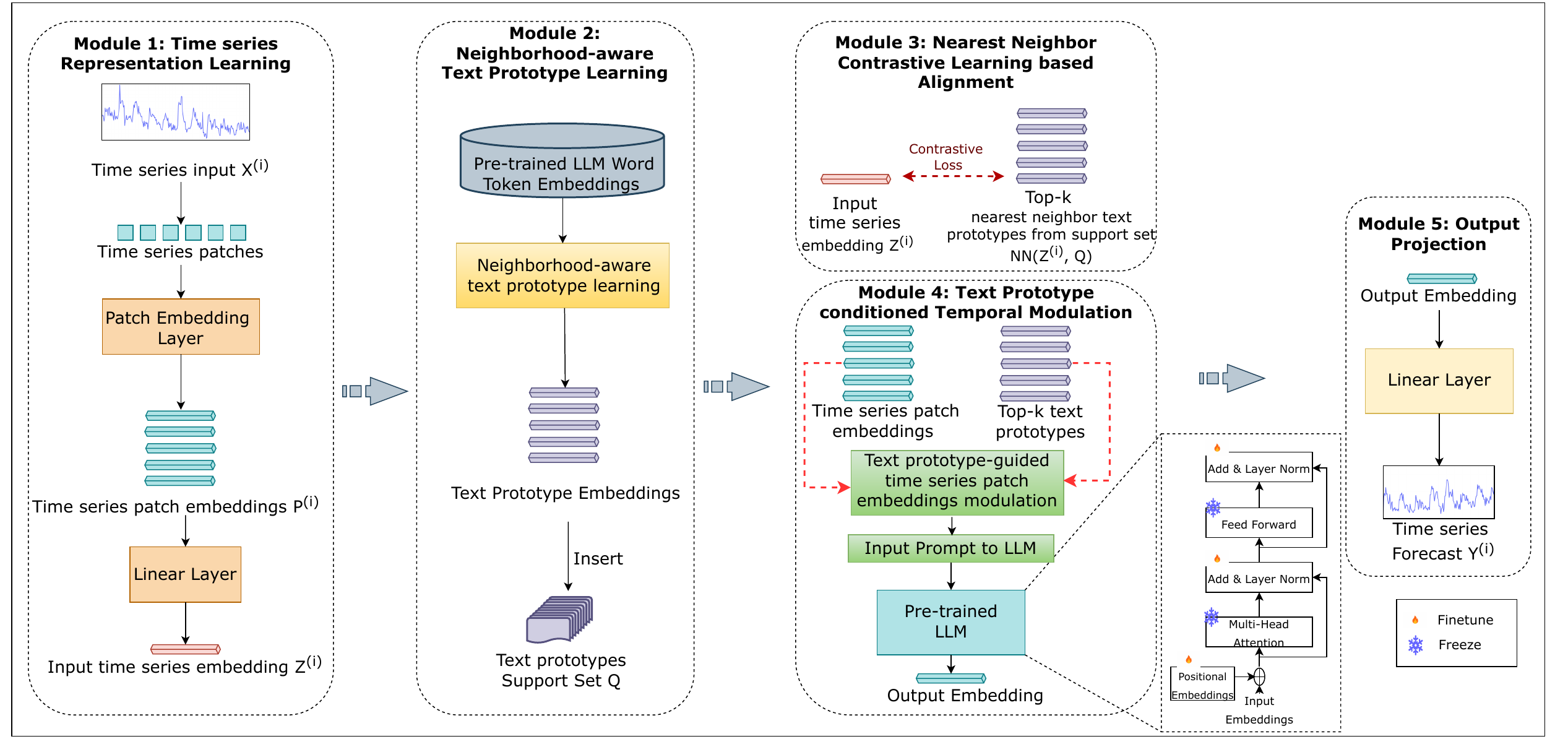}
  \caption{Illustration of NeST architecture. Given the univariate time series input, we divide it into patches of local time windows and obtain the time series embedding $Z^{(i)}$ using a patch embedding layer and a linear layer (Module 1). Pre-trained word token embeddings of the LLM are used to generate neighborhood aware text prototypes (Module 2). Each learned text prototype can be considered as a representative anchor of the each local word token embedding region. We use NNCL by inserting the text prototypes into a support set at the end of each training step. The top-$k$ nearest-neighbor text prototypes from the support set is retrieved and we compute the contrastive loss; ($L_{\text{NNCL}}$) between the time series embedding and the top-$k$ nearest-neighbor text prototypes (Module 3). The input prompt to finetune the LLM is formulated using the text prototype conditioned temporal modulation (Module 4). Only the layer normalization and positional embeddings in the LLM are finetuned. Output prediction of the LLM is flattened and then followed by a linear layer to get the final forecast; \(\hat{Y}^{(i)}\) (Module 5).}
  \label{fig:method}
\end{figure*}

\section{Related Work} 
\subsection{Time Series Forecasting} 
Time series forecasting has been extensively studied using a broad spectrum of approaches, ranging from traditional statistical methods to modern deep learning architectures. Classical forecasting methods include autoregressive integrated moving average (ARIMA), seasonal ARIMA (SARIMA), vector autoregressive, and exponential smoothing models \cite{makridakis2018m4}. More recently, deep neural networks (DNNs) have demonstrated strong predictive performance across diverse forecasting tasks. These methods encompass recurrent neural networks (RNNs) \cite{lai2018modeling,smyl2020hybrid}, temporal convolutional networks (TCNs) \cite{sen2019think,wu2022timesnet}, transformer-based architectures \cite{nie2022time,wu2021autoformer,zhou2021informer,liu2023itransformer}, multilayer perceptron (MLP)-based models \cite{zeng2023transformers,oreshkin2019n,challu2023nhits}, and graph neural networks (GNNs) \cite{wu2020connecting}. Despite their success, most existing forecasting models are designed for specific domains with a specific set of data characteristics, limiting their ability to generalize across heterogeneous datasets and forecasting scenarios \cite{jiang2024empowering}. 

\subsection{Large Language Models for Time Series Forecasting} 

The emergence of LLMs, including LLaMA \cite{touvron2023llama}, GPT2 \cite{radford2019language}, and T5 \cite{raffel2020exploring}, has motivated growing interest in extending their capabilities beyond natural language processing to other modalities such as time series \cite{jiang2024empowering,jin2024position}, video \cite{ataallah2024minigpt4}, and tabular data \cite{hegselmann2023tabllm}. Among these directions, adapting LLMs for time series forecasting has attracted significant attention owing to the strong prior knowledge encoded in pretrained LLMs and the sequential nature of both text and time series. One of the earliest studies, PromptCast \cite{xue2023promptcast}, transformed numerical time series into natural language descriptions and finetuned an LLM using a large prompt based corpus. However, this approach required extensive manual generation of textual annotations. To alleviate this limitation, LLMTime \cite{gruver2024large} proposed a specialized tokenization strategy that enables LLMs to directly process numerical sequences, reducing the dependence on textual annotations. Subsequent work explored alternative adaptation mechanisms. For instance, Zhou et al. \cite{zhou2023one} segmented time series into patches inspired by PatchTST \cite{nie2022time} and finetuned LLMs using patch sequences. In contrast, Time-LLM \cite{jin2023time} and TEST \cite{sun2023test} employed reprogramming strategies to align time series representations with the language space, enabling the utilization of pretrained LLMs with minimal parameter updates. Recent methods have further highlighted the importance of prompt design and representation alignment. TEMPO \cite{cao2023tempo} and $S^{2}$IP-LLM \cite{pan2024s} demonstrated that text prototypes derived from LLM word embeddings through linear probing can effectively bridge the gap between temporal and linguistic representations. Both approaches rely on seasonal-trend decomposition using Loess (STL) \cite{cleveland1990stl} to construct time series embeddings. However, STL-based decomposition can struggle to accurately capture seasonal patterns in the presence of temporal shifts and non-stationary fluctuations, which frequently occur in real-world time series \cite{wen2019robuststl,phungtua2024adaptive}. Similarly, S$^2$TS-LLM \cite{s2tsllm} leverages time-frequency transformations to extract informative representations from time series data. While effective, such transformations may compress important temporal dynamics, particularly when forecasting from short historical sequences. Despite the current advances, existing LLM based forecasting frameworks largely rely on linear probing, handcrafted decomposition techniques, or computationally intensive prompt formulation mechanisms. In contrast, we introduce a neighborhood-aware text prototype learning framework that learns informative text prototypes directly from LLM word embeddings without requiring STL decomposition or other specialized pre-processing techniques. In addition, we propose a nearest-neighbor contrastive learning (NNCL) objective to enhance the alignment between temporal and textual representations. To the best of our knowledge, we are the first to adopt NNCL to harness the power of LLMs for time series forecasting. Furthermore, we incorporate the learned prototypes with time series to formulate the prompt through a new text prototype conditioned temporal modulation module, avoiding the limitations of simple feature concatenation and computationally expensive attention based fusion methods.

\section{Methodology}

This section presents the proposed NeST method, illustrated in Fig. \ref{fig:method}. NeST consists of four main modules. First, in the time series representation learning module, we partition the multivariate time series into univariate time series and process them independently following a channel independence strategy \cite{nie2022time}. The univariate time series is divided into patches and a mapping function is used to obtain the embedding of the univariate time series. In the second module, we introduce a neighborhood-aware mapping function that leverages local relationships to learn a set of text prototypes from the word token embeddings of the LLM. The third module aligns the learned text prototypes with the time series representations through a nearest-neighbor contrastive learning (NNCL) objective, thereby encouraging consistency between textual and time series representations. To construct the prompt for the LLM, the fourth module employs a text prototype conditioned modulation mechanism that integrates the learned prototypes into the time series patch embeddings, enabling the prototypes to guide representation learning. Since prior studies have shown that the self-attention and feed forward layers of LLMs retain the majority of the knowledge acquired during pretraining \cite{zhou2023one,zhao2023tuning}, NeST adapts the LLM for time series forecasting by finetuning only the layer normalization and positional embedding parameters while preserving the learned prior of the LLM.

\subsection{Problem Definition}

We define the time series forecasting problem as follows: Let \( X \in \mathbb{R}^{M \times T} \) be the input multivariate time series of historical observations where \(M\) is the number of variables or channels and \(T\) is the time steps. Since \(X\) consists of \(M\) channels, \(X^{(i)} \in \mathbb{R}^{1 \times T} \) represents the univariate time series corresponding to channel \(i\). Given \(X^{(i)}\), our objective is to predict the data points for \(H\) future time steps as denoted by \(\hat{Y}^{(i)} \in \mathbb{R}^{1 \times H} \) while minimizing the error between the prediction \(\hat{Y}^{(i)}\) and the ground truth \(Y^{(i)} \in \mathbb{R}^{1 \times H}\).

\subsection{Time Series Representation Learning}
\label{sec:method.timerepresention}

The input multivariate time series \(X\) is partitioned into \(M\) univariate time series and \(X^{(i)}\) denotes data points from one channel, (\(i\)). Afterwards, each univariate time series is processed independently. The input time series \(X^{(i)}\) is normalized using reversible instance normalization (RevIN) \cite{kim2021reversible} to avoid the effects of temporal distribution shift. The concept of time series patching is adopted to obtain the time series embeddings as it allows to capture relevant information from the time series with a long look-back window while minimizing computational costs \cite{nie2022time}. In the time series patching operation, the normalized time series \(X^{(i)}\) is divided into patches where the length of a patch is \(C\) and the stride is \(S\). After the patching operation, the total number of patches created is defined by \(N = [(T - C)/ S] + 2 \). A 1-Dimensional (1D) convolution layer is utilized to obtain the time series patch embeddings as \(P^{(i)} \in \mathbb{R}^{N \times D}\), where the dimension of the embedding is denoted by \(N \times D\). The patch embeddings are finally fed into a simple linear layer to generate the embedding of the input univariate time series; \(Z^{(i)} \in \mathbb{R}^{1 \times D}\). 

\subsection{Neighborhood aware Text Prototype Learning}
\label{sec:method.neighborhoodawaretexttrototype}

\subsubsection{Motivation:} Let the word token embeddings in the LLM be denoted as \(W \in \mathbb{R}^{V \times D}\), where \(V\) is the size of the vocabulary and \(D\) represents the embedding dimension of the pretrained LLM. Most LLMs typically contain approximately 50000 word tokens \cite{radford2019language} in the vocabulary. Consequently, directly incorporating the full set of token embeddings into prompt construction introduces substantial computational overhead \cite{jin2023time,pan2024s,cao2023tempo,sun2023test}. To alleviate this issue, existing approaches typically employ a learnable linear projection layer to generate a compact set of text prototypes. The parameters of this projection layer are learned solely by minimizing the forecasting loss. Therefore, the generated prototypes are optimized to improve prediction accuracy without any explicit constraint to preserve the structural properties of the pretrained token embedding space. The pretrained word token embedding space exhibits meaningful geometric relationships, where semantically related tokens tend to form local neighborhoods and occupy nearby regions, while unrelated tokens are positioned further apart. These neighborhood structures are encoded during large scale LLM pretraining and capture useful semantic features among tokens. However, since existing projection based methods do not explicitly preserve these local geometric relationships, the resulting prototypes may be mapped to regions that are sparsely populated by valid token embeddings or lie outside the manifold formed by the pretrained word token embeddings, as illustrated by Fig. \ref{fig:textprototypesvis}. As a result, the generated prototypes may be less compatible with the representations expected by the pretrained LLM. This limitation is particularly problematic in few-shot forecasting settings, where only a small number of training samples are available. In such cases, the projection layer can easily overfit the limited observations and learn task-specific representations that limit generalization capabilities, and reducing forecasting robustness. Motivated by these limitations, we propose a neighborhood-aware text prototype learning mechanism that explicitly preserves local neighborhood relationships within the pretrained token embedding space. Specifically, each learned text prototype serves as a representative anchor of each local token embedding region.

\subsubsection{Design:} We introduce a set of learnable text prototypes; \(E \in \mathbb{R}^{U \times D}\), where \(U\) (\(U<<V\)) is the number of text prototypes. The prototypes are initialized randomly and subsequently optimized during training. The prototype updates are driven by the distribution of word token embeddings in the latent space, allowing them to adapt dynamically and align with the underlying structure of the token embedding manifold. Each \(i^{th}\) frozen word token embedding; \(w_i\) is assigned to its nearest text prototype; \(e^{*}_i\) in Euclidean distance as denoted by Eq.\ref{eq:method.proto}:
\begin{equation}
\label{eq:method.proto}
e^{*}_i = \argmin _{e_j \in E} \| w_i - e_j \|_2
\end{equation}
The text prototypes are optimized by minimizing the mean squared error (MSE) loss between the word token and their nearest text prototype embeddings is calculated as \(\mathcal{L}_{proto}\) using Eq. \ref{eq:method.lproto}.
\begin{equation}
\label{eq:method.lproto}
\mathcal{L}_{\text{proto}} = \frac{1}{V} \sum\limits_{i=1}^V  \norm{w_{i} - e^{*}_{i}}_2^{2}
\end{equation}
This encourages each prototype to act as a centroid of a dense region in the LLM word token embedding space, preserving its geometric and semantic priors while significantly reducing the dimensionality.

\subsection{Nearest-Neighbor Contrastive Learning-based Cross-modal Alignment}

\subsubsection{Motivation}

While neighborhood-aware text prototype learning ensures that the text prototypes are representative of the word token embedding space, these prototypes must also align with the time series representations. To achieve this, we adopt Nearest-Neighbor Contrastive Learning (NNCL) \cite{dwibedi2021little}. Conventional contrastive learning is built upon the idea of learning a representation space in which positive pairs (e.g. augmentations of the same time series) are pulled closer, while the negative samples (e.g. other time series in the batch) are pushed apart \cite{chen2020simple}. On the other hand, neighboring samples are more likely to lie on the same underlying semantic manifold and therefore provide a more informative approximation of the data distribution. This encourages locally smooth representations and reduces sensitivity to noise or imperfect augmentations. Furthermore, by retrieving nearest-neighbor text prototypes from a large support set, the model can leverage a richer set of semantically related representations than those available within a mini-batch. Such neighborhood-based supervision has been shown to improve representation quality, robustness, and generalization, particularly in few-shot learning scenarios where learning meaningful cross-modal alignments from limited data is challenging.

\subsubsection{Design:} We maintain a first in first out (FIFO) support set; \(Q \in \mathbb{R}^{q \times D}\), where \(q\) (\(q>>U\)) is the length of the queue and \(D\) is the embedding dimension of text prototypes. After each training step, we update the support set by inserting the current batch of text prototypes ($E$) to the end of the queue and removing the oldest batch. By formulating this support set that has a large enough capacity, we aim to approximate and represent the full text prototype data distribution in the embedding space which can also be useful for few-shot learning. Positive pairs are formed by utilizing the nearest neighbors of the time series embedding $Z^{(i)}$ in the support set; $Q$ and the loss function is defined as Eq. \ref{eq:nnloss}. The contrastive loss: $L_\text{NNCL}$ is used to align the text prototypes with the time series, ensuring that the text prototypes are representative of the time series characteristics.
\begin{equation}
\label{eq:nnloss}
\mathcal{L}_\text{NNCL} = -  \log{
	{
		\frac{\exp{(\text{NN}{_k}(Z^{(i)}, Q)\cdot Z^{(i)}/\tau)}}
		{ \sum\limits_{b=1}^B\exp{(\text{NN}{_k}(Z^{(i)}, Q)\cdot Z^{(i)}_b / \tau)}}
	}
}
\end{equation}
In Eq. \ref{eq:nnloss}, NN$_k$\((Z^{(i)}, Q)\) is the operator to obtain the top-\(k\) nearest-neighbor text prototypes from the support set as denoted by Eq. \ref{eq:nn} and $\tau$ and \(B\) represent the softmax temperature and batch size respectively. All embeddings stored in the queue are detached from gradient updates and are $l_2$ normalized before nearest-neighbor retrieval.

\begin{equation}
\label{eq:nn}
\operatorname{NN}_k\!\left(Z^{(i)}, \mathcal{Q}\right)
=
\operatorname*{arg\,min}_{k,\; Q^{(j)} \in \mathcal{Q}}
\left\|Z^{(i)} - Q^{(j)}\right\|_2
\end{equation}

\subsection{Prompt Formulation via Text Prototype Conditioned Temporal Modulation}
 
\subsubsection{Motivation:} Existing methods typically incorporate textual information to the time series representations through token concatenation, prompt prefixing, or cross attention based fusion. However, these strategies directly modify the token sequence, resulting in strong entanglement between semantic priors and temporal representations. Such tight coupling may disrupt the intrinsic temporal structure encoded at the time series patch level embeddings and offers limited control over how textual guidance influences time series features. To address this limitation, we propose text prototype conditioned temporal modulation, a feature-wise conditioning mechanism that integrates textual semantics via learned modulation of temporal embeddings. Instead of modifying the token sequence or introducing auxiliary text tokens, this mechanism leverages text prototypes to guide temporal representations through continuous and parameterized feature modulation.

\begin{figure}
  \centering
  \includegraphics[width=\linewidth]{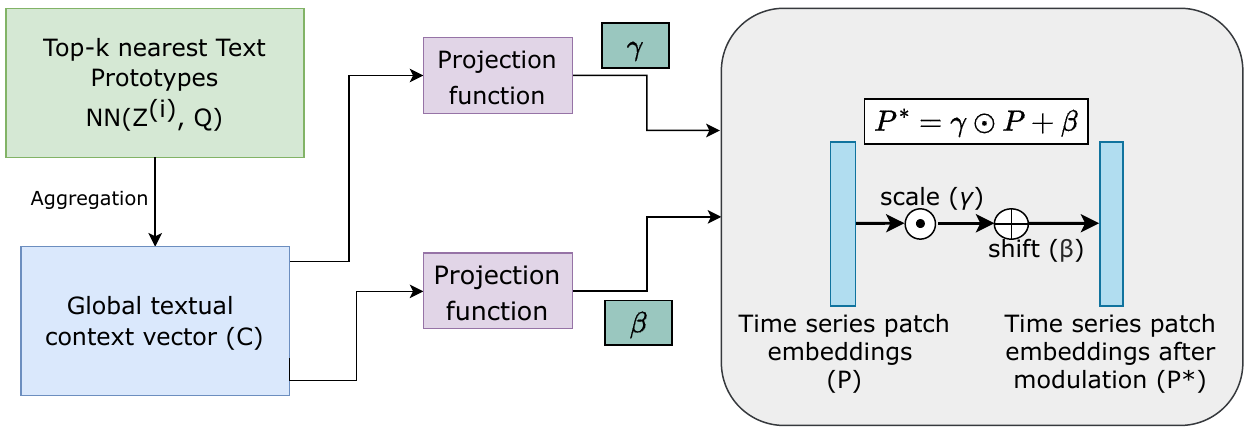}
  \caption{Mechanism of text prototype conditioned temporal modulation. Top-k nearest text prototypes are aggregated into a global textual context vector (C). C is used to compute $\gamma$ and $\beta$, which are feature-wise modulation parameters learned using a projection function. Time series patch embeddings after the modulation ($P^*$) is used as the input prompt to the LLM.}
  \label{fig:modulation}
\end{figure}

\subsubsection{Design:} As demonstrated in Fig. \ref{fig:modulation}, we first aggregate the top-k text prototypes (NN \((Z^{(i)}, Q)\)) to obtain a global textual context vector (\(C\)). The aggregated context vector is then used to compute feature-wise modulation parameters as denoted by Eq. \ref{eq:method.context}:
\begin{equation}
\label{eq:method.context}
(\gamma, \beta) = f(C)
\end{equation}
where $\gamma, \beta \in \mathbb{R}^{1 \times D}$, and $f(\cdot)$ is a lightweight learnable projection function (e.g. linear layer). Text prototype guided temporal modulation is applied through a residual interpolation mechanism as denoted by Eq. \ref{eq:method.modulation}:
\begin{equation}
\label{eq:method.modulation}
P^{(i)*} = \gamma \odot P^{(i)} + \beta
\end{equation}
where $\odot$ denotes element wise multiplication and $P$ is the time series patch embeddings. $P^{(i)*}$ is used as the input prompt to finetune the LLM.

\subsection{LLM finetuning and Output projection}
We follow a finetuning strategy \cite{zhou2023one} where only the layer normalization and positional embeddings in the LLM are finetuned while keeping the multi-head attention and feed-forward layers intact. Finetuning of the LLM can enable the capability to forecast the time series with the help of the formulated input prompt while preserving the prior it has learned during extensive training. Output prediction of the LLM is flattened and then followed by a linear layer to get the final forecast; \(\hat{Y}^{(i)}\). We compute the MSE loss between the forecast \(\hat{Y}^{(i)}\) and the ground truth \(Y^{(i)}\) using Eq. \ref{eq.lforecast}.
\begin{equation}
\label{eq.lforecast}
\mathcal{L}_{\text{forecast}} = \frac{1}{B} \sum\limits_{b=1}^B \norm{\hat{Y}^{(i)}_b - Y^{(i)}_b}_2^{2}
\end{equation}

\subsection{Training}

\textbf{Overall Optimization Objective.} The total loss used in our framework is shown in Eq. \ref{eq:totalloss}, where $\lambda$ ($> 0$) is the weight which controls the contribution of the components in the loss. This loss function is proposed considering all the components in NeST. This ensures that the learned embedding space and the formulated prompt implicitly align across the two modalities: text and time series, while optimally leveraging the rich representation space learned by the LLMs to generate text prototypes via end-to-end finetuning.
\begin{equation}
\label{eq:totalloss}
    \mathcal{L}_{\text{total}} = \mathcal{L}_{\text{forecast}} + \lambda (\mathcal{L}_\text{NNCL} + \mathcal{L}_{\text{proto}}) 
\end{equation}

\section{Experiments}\label{sec2}

\subsection{Datasets}
We evaluate the proposed method on a comprehensive collection of publicly available time series datasets spanning across diverse application domains, including energy, electricity, transportation, meteorology, finance, and economics. The detailed statistics are presented in Table \ref{tab:exp.datasetstats}, including information about dimension, time series lengths, dataset sizes, sampling frequency, and source domains. The experimental results are reported for long‑term, short‑term, few‑shot, and zero‑shot forecasting settings. In addition to standard benchmarks, we further assess performance on real‑world distributed photovoltaic (PV) power forecasting datasets to demonstrate the practical applicability of our method.

\begin{table*}[t] 
\centering 
\fontsize{8pt}{8pt}\selectfont 
\centering 
\caption{Statistics of the datasets \cite{wu2022timesnet}. Dimensions of the datasets are included as Dim which indicates the number of channels in the time series. The dataset size includes the training, validation and testing splits.}
\label{tab:exp.datasetstats} 
\begin{tabular}{ccccccc} 
\toprule[1.0pt] 
{Tasks}   & {Dataset} & {Dim} & {Series Length} & {Dataset Size} & {Frequency} & {Domain} \\  
\midrule[1.0pt] 
\multirow{7}{*}{Long-term Forecasting}
& 	ETTh1	 & 	7	 & 	[96, 192, 336, 720] & (8545, 2881, 2881)    & 	1 hour 	 &  Temperature	  \\
\cmidrule{2-7}
& 	ETTh2	 & 	7	 & 	[96, 192, 336, 720] & (8545, 2881, 2881)    & 	1 hour 	 &  Temperature	  \\
\cmidrule{2-7}
& 	ETTm1	 & 	7	 & 	[96, 192, 336, 720] & (34465, 11521, 11521)    & 	15 min 	 &  Temperature	  \\
\cmidrule{2-7}
& 	ETTm2	 & 	7	 & 	[96, 192, 336, 720] & (34465, 11521, 11521)    & 	15 min 	 &  Temperature	  \\
\cmidrule{2-7}
& 	Weather	 & 	21	 & 	[96, 192, 336, 720] & (36792, 5271, 10540)    & 	10 min 	 &  Weather	  \\
\cmidrule{2-7}
& 	Traffic	 & 	862	 & 	[96, 192, 336, 720] & (12185, 1757, 3509)    & 	1 hour 	 &  Transportation	  \\
\cmidrule{2-7}
& 	ECL	 & 	321	 & 	[96, 192, 336, 720] & (18317, 2633, 5261)    & 	1 hour 	 &  Electricity	  \\
\midrule[0.5pt] 
\multirow{6}{*}{Short-term Forecasting}
& 	M4-Yearly	 & 	1	 & 	6 & (23000, 0, 23000)    & 	Yearly 	 &  Demographic	  \\
\cmidrule{2-7}
& 	M4-Quarterly	 & 	1	 & 	8 & (24000, 0, 24000)    & 	Quarterly 	 &  Finance	  \\
\cmidrule{2-7}
& 	M4-Monthly	 & 	1	 & 	18 & (48000, 0, 48000)    & 	Monthly 	 &  Industry	  \\
\cmidrule{2-7}
& 	M4-Weekly	 & 	1	 & 	13 & (359, 0, 359)    & 	Weekly 	 &  Macro	  \\
\cmidrule{2-7}
& 	M4-Daily	 & 	1	 & 	14 & (4227, 0, 4227)    & 	Daily 	 &  Micro	  \\
\cmidrule{2-7}
& 	M4-Hourly	 & 	1	 & 	48 & (414, 0, 414)    & 	Hourly 	 &  Other	  \\
\bottomrule[1.0pt] 
\end{tabular} 
\end{table*} 

\begin{table*}[t]
\centering
\setlength{\tabcolsep}{2pt}
\caption{Long-term forecasting results. The look-back window length is set to 512. The average MSE and MAE values across the four prediction horizons; $H \in$ [96, 192, 336, 720] are reported. Lower values indicate better performance. Bold: the best, Underline: the second best.}
\label{tab:exp.longtermforecasting}
\fontsize{8pt}{8pt}\selectfont
\centering
\begin{tabular}{ccccccccc}
\toprule[1.0pt]
{Method}   & {NeST (Ours)} & {$S^{2}$TS-LLM} & {$S^{2}$IP-LLM} & {Time-LLM} & {OFA} &  {PatchTST}  & TimesNet & {DLinear} \\
\cmidrule(lr){1-1}
\cmidrule(lr){2-2}
\cmidrule(lr){3-3} 
\cmidrule(lr){4-4}
\cmidrule(lr){5-5}
\cmidrule(lr){6-6}
\cmidrule(lr){7-7}
\cmidrule(lr){8-8}
\cmidrule(lr){9-9}
{Metric}          & MSE~~MAE               & MSE~~MAE                    & MSE~~MAE          & MSE~~MAE          & MSE~~MAE    & MSE~~MAE   & MSE~~MAE      & MSE~~MAE  \\
\midrule[0.5pt]
	 ECL	 & 	\textbf{0.160}~~\textbf{0.254}	 & 	\underline{0.164}~~\underline{0.256} & 0.166~~0.262 	 & 	0.167~~0.264 	 & 	0.167~~0.263 	 & 	0.216~~0.304 	  	 & 	0.192~~0.295 & 0.166~~0.263	 \\
\midrule[0.5pt]
	 Traffic	 & 	\textbf{0.392}~~\textbf{0.270}	 & 	\underline{0.396}~~\underline{0.278} & 0.405~~0.286 & 0.407~~0.289 & 0.414~~0.294 	 & 	0.481~~0.304  & 0.620~~0.336 & 0.433~~0.295	  \\
\midrule[0.5pt]
	 Weather	 & 	\underline{0.226}~~\textbf{0.263} 	 & 	\textbf{0.225}~~\underline{0.265}  & 0.228~~0.265 	 & 	0.237~~0.269 	 & 	0.237~~0.270 	 & 0.259~~0.281  		 & 	0.259~~0.287 & 0.248~~0.300\\
\midrule[0.5pt]
	 {ETTh$_1$}	 & \textbf{0.395}~~\textbf{0.422}	& \underline{0.404}~~\underline{0.429}  & 0.418~~0.436 & 0.426~~0.435 & 0.427~~0.426 & 0.469~~0.454 & 0.458~~0.450 & 0.418~~0.439 \\
\midrule[0.5pt]
{ETTh$_2$} &    \textbf{0.319}~~\textbf{0.374} & \underline{0.321}~~\underline{0.376} & 0.355~~0.399 &    0.361~~0.398 &   0.354~~0.394     & 0.387~~0.407   & 0.414~~0.427 & 0.502~~0.481 \\
\midrule[0.5pt]
{ETTm$_1$} &    \textbf{0.344}~~\textbf{0.376}  & \underline{0.344}~~\underline{0.380}   &   0.346~~0.382     &   0.354~~0.384     &   0.352~~0.383 &   0.387~~0.400       &   0.400~~0.406    & 0.357~~0.389 \\
\midrule[0.5pt]
{ETTm$_2$} & \textbf{0.251}~~\textbf{0.314} &   \underline{0.258}~~\underline{0.318}     &   0.262~~0.326     &   0.275~~ 0.334     &   0.266~~ 0.326   & 0.281 ~~0.326         & 0.291~~0.333  & 0.275 ~~0.340 \\
\bottomrule[1.0pt]
\end{tabular}
\end{table*}

\begin{table*}[t]
\centering
\setlength{\tabcolsep}{2pt}
\caption{Few-shot forecasting results. Only $10\%$ of the training data is used for finetuning. The look-back window length is set to 512. The average MSE and MAE values across the four prediction horizons; $H \in$ [96, 192, 336, 720] are reported. Lower values indicate better performance. Bold: the best, Underline: the second best.}
\label{tab:exp.fewshotforecasting}
\fontsize{8pt}{8pt}\selectfont
\centering
\begin{tabular}{ccccccccc}
\toprule[1.0pt]
{Method}   & {NeST (Ours)} & {$S^{2}$TS-LLM} & {$S^{2}$IP-LLM} & {Time-LLM} & {OFA} &  {PatchTST}  & TimesNet & {DLinear} \\
\midrule[0.5pt]
{Metric}          & MSE~~MAE               & MSE~~MAE                    & MSE~~MAE          & MSE~~MAE          & MSE~~MAE    & MSE~~MAE   & MSE~~MAE      & MSE~~MAE   \\
\midrule[1.0pt]
	 {ETTh$_1$}	 & \textbf{0.536}~~\underline{0.507}  &   \underline{0.543}~~\textbf{0.493}    & 0.593~~0.529  & 0.785~~0.553  &   0.590~~0.525    &   0.633~~0.542      &   0.869~~0.628    & 0.691~~0.600 \\
\midrule
    {ETTh$_2$}	 &   \textbf{0.363}~~\underline{0.404}  &   \underline{0.372}~~\textbf{0.402}    &   0.419~~0.439    &   0.424~~0.441    &   0.397~~0.421    & 0.415~~0.431      &   0.479~~0.465    &   0.605~~0.538 \\
\midrule
{ETTm$_1$}	 &   \underline{0.417}	~~0.436    & 0.465~~0.436 &   0.455~~\underline{0.435}    &   0.487~~0.461    &   0.464~~0.441    &   0.501~~0.466       &   0.677~~0.537    &   \textbf{0.414}~~\textbf{0.428} \\
\midrule
{ETTm$_2$}  &   \textbf{0.277}~~\textbf{0.326}  &   \underline{0.279}~~\underline{0.327}    & 0.284 ~~0.332     &   0.305 ~~0.344   &   0.293 ~~0.335   &   0.296 ~~0.343      &   0.320 ~~0.353   & 0.298~~0.352 \\

\bottomrule[1.0pt]
\end{tabular}
\end{table*}

\subsubsection{Long-term Forecasting Benchmark Datasets}
The performance of long-term forecasting task is evaluated on seven widely used benchmark datasets introduced in prior work \cite{zhou2021informer}, including Electricity Transformer Temperature (ETT), Weather, Traffic, and Electricity Consumption Load (ECL). The ETT datasets consist of electricity load and oil temperature measurements from power transformers, collected over a period of two years from two locations in China. Based on the sampling frequency, the ETT datasets are categorized into ETTh$_1$, ETTh$_2$ (hourly sampling) and ETTm$_1$, ETTm$_2$ (15‑minute sampling). The weather dataset contains one year of meteorological observations from 21 weather stations in Germany, sampled every 10 minutes. The traffic dataset comprises hourly road occupancy measurements from 862 freeway sensors deployed across California. The ECL dataset records hourly electricity consumption from 321 customers. Few-shot and zero-shot forecasting performance is evaluated using the ETT datasets; ETTh$_1$, ETTh$_2$, ETTm$_1$, and ETTm$_2$.

\subsubsection{Short-term Forecasting Benchmark Datasets}
Short‑term forecasting performance is evaluated on the M4 dataset \cite{makridakis2018m4}, a large‑scale collection of 100,000 univariate time series from heterogeneous real-world domains commonly used in business, economic, and financial forecasting. The M4 dataset is subdivided into six subsets according to sampling frequency, including hourly, daily, weekly, monthly, quarterly, and yearly series.

\subsubsection{Distributed Photovoltaic (PV) Power Forecasting Datasets}
To assess the real‑world applicability of our method, we further evaluate on nine distributed PV power generation datasets collected from two publicly available studies \cite{yao2021photovoltaic, chen2022solar}. These datasets represent PV power generation outputs originated from different geographically distributed stations under varying operating and environmental conditions. Specifically, there are five datasets (Stations 1, 2, 4, 7, and 8) from one source \cite{yao2021photovoltaic}, and there are four datasets  (Stations 1, 2, 5, and 6) obtained from another source \cite{chen2022solar}. For each dataset, we used 70\%, 5\% and 25\% splits for training, validation, and testing respectively. 

\subsection{Baselines}
We compare our method with state-of-the-art (SOTA) methods in time series forecasting. The baseline methods encompass LLM-based time series models; S\textsuperscript{2}TS-LLM \cite{s2tsllm}, S\textsuperscript{2}IP-LLM \cite{pan2024s}, Time-LLM \cite{jin2023time} and OFA \cite{zhou2023one} and a transformer-based time series model; PatchTST \cite{nie2022time}, a CNN-based time series model; TimesNet \cite{wu2022timesnet}, and a MLP-based time series model; DLinear \cite{zeng2023transformers}. In our comparative analysis of short-term forecasting task, we additionally compare with N-HiTS \cite{challu2023nhits}. Furthermore, in the distributed PV power forecasting experiments, we compare the proposed method with solar forecasting specific baselines, encompassing LLM-based (RAG4PVF) \cite{gao2026pre}, transformer‑based \cite{kim2024multi}, and LSTM‑based \cite{wang2019photovoltaic} methods which are tailored specifically for PV generation forecasting modeling. We further include strong generic forecasting methods commonly adopted in energy applications, including OFA \cite{zhou2023one}, PatchTST \cite{nie2022time}, DLinear \cite{zeng2023transformers}, and Informer \cite{zhou2021informer}.

\subsubsection{Evaluation Metrics}
We use mean square error (MSE) and mean absolute error (MAE) to evaluate the performance of the model in terms of long-term, few-shot, and zero-shot forecasting. To evaluate the short-term forecasting performance on the M4 benchmark dataset, we utilize symmetric mean absolute percentage error (SMAPE), mean absolute scaled error (MASE) and overall weighted average (OWA) \cite{oreshkin2019n}. The OWA metric was specifically introduced for the evaluation of short-term forecasting performance on M4 dataset. To evaluate the performance of distributed PV power forecasting, we use Root Mean Square Error (RMSE) and Coefficient of Determination (R$^2$ score) \cite{gao2026pre}. The calculations of the mentioned metrics are presented as follows:
\begin{align*}
\label{equ:metrics}
    \text{MSE} &= \frac{1}{H}\sum_{h=1}^T (\mathbf{Y}_{h} - \Hat{\mathbf{Y}}_{h})^2 \\
    \text{MAE} &= \frac{1}{H}\sum_{h=1}^H|\mathbf{Y}_{h} - \Hat{\mathbf{Y}}_{h}|\\
    \text{SMAPE} &= \frac{200}{H} \sum_{h=1}^H \frac{|\mathbf{Y}_{h} - \Hat{\mathbf{Y}}_{h}|}{|\mathbf{Y}_{h}| + |\Hat{\mathbf{Y}}_{h}|} \\
    \text{MAPE} &= \frac{100}{H} \sum_{h=1}^H \frac{|\mathbf{Y}_{h} - \Hat{\mathbf{Y}}_{h}|}{|\mathbf{Y}_{h}|} \\
    \text{MASE} &= \frac{1}{H} \sum_{h=1}^H \frac{|\mathbf{Y}_{h} - \Hat{\mathbf{Y}}_{h}|}{\frac{1}{H-s}\sum_{j=s+1}^{H}|\mathbf{Y}_j - \mathbf{Y}_{j-s}|} \\
    \text{OWA} &= \frac{1}{2} \left[ \frac{\text{SMAPE}}{\text{SMAPE}_{\textrm{Naïve2}}}  + \frac{\text{MASE}}{\text{MASE}_{\textrm{Naïve2}}}  \right] \\
    \text{RMSE} &= \sqrt{\frac{1}{H}\sum_{h=1}^T (\mathbf{Y}_{h} - \Hat{\mathbf{Y}}_{h})^2} \\
    \text{R$^2$} &= 1 - \frac{\sum_{h=1}^{H} (\mathbf{Y}_{h} - \hat{\mathbf{Y}}_{h})^2} {\sum_{h=1}^{H} (\mathbf{Y}_{h} - \bar{\mathbf{Y}})^2}
\end{align*}

where $s$ is the periodicity of the time series and $H$ is the prediction interval or the prediction horizon. The $h^{th}$ ground truth time series and prediction forecast are denoted as $Y_{h}$ and $\hat{Y}_{h}$ respectively where $h \in {\{1, ... , H\}}$.

\subsection{Implementation Details}
We use GPT2 \cite{radford2019language} as the pretrained LLM backbone in our framework and existing LLM-based methods. During fine‑tuning, we update the layer normalization and positional embedding parameters in the first six Transformer blocks, while keeping the remaining model weights frozen \cite{jin2023time, pan2024s}. All models are implemented in PyTorch \cite{paszke2019pytorch} and trained on NVIDIA A100 GPUs (80 GB memory). We evaluate the impact of the number of text prototypes by varying it in [1000, 5000, 8000]. The corresponding support set sizes are scaled proportionally, set to [$\times20, \times4, \times3$] times the number of text prototypes, respectively. Number of text prototypes was empirically determined as 1000 for ETTh$_1$, ETTh$_2$, M4 and distributed PV forecasting datasets, 5000 for ETTm$_1$ and ETTm$_2$ datasets, and 8000 for traffic, ECL, and weather datasets. The number of top‑k nearest‑neighbor text prototypes is empirically determined as k=8 for all experiments. The embedding dimensionality is set to 768, matching the token embedding dimension of GPT2 model. For long‑term forecasting, the input look‑back window length is fixed to 512, and evaluation is performed over four forecasting horizons; [96, 192, 336, 720]. Reported results are averaged across these horizons. In long-term forecasting task, we divide the time series input into patches of length 48 and stride 48. For few‑shot forecasting, we conduct long‑term forecasting experiments using only 10\% of the training data for finetuning. For zero‑shot forecasting, NeST is evaluated in a cross‑domain setting, where the model is trained on a source dataset and directly tested on an unseen target dataset without task‑specific adaptation. The look‑back window length, prediction horizons, and evaluation metrics follow the same configuration as in long‑term forecasting. For short‑term forecasting, prediction horizons range from 6 to 48, and the input sequence length is set to twice the prediction horizon, following standard practice. In distributed PV power forecasting experiments, the look‑back window length is set to 48, with a prediction horizon of 24, corresponding to forecasting the next 24 hours of PV power generation based on the previous 48 hours of historical observations.

\begin{table*}[t]
\centering
\setlength{\tabcolsep}{2pt}
\caption{Zero-shot forecasting results. The model is trained on a source dataset and
directly tested on an unseen target dataset, which is indicated by source $\rightarrow$ target. The look-back window length is set to 512. The average MSE and MAE values across the four prediction horizons; $H \in$ [96, 192, 336, 720] are reported. Lower values indicate better performance. Bold: the best, Underline: the second best.}
\label{tab:exp.zeroshotforecasting}
\fontsize{7pt}{7pt}\selectfont
\centering
\begin{tabular}{ccccccccc}
\toprule[1.0pt]
{Method}   & {NeST (Ours)} & {$S^{2}$TS-LLM} & {$S^{2}$IP-LLM} & {Time-LLM} & {OFA} &  {PatchTST} & TimesNet & {RLinear} \\
\midrule[0.5pt]
{Metric}          & MSE~~MAE               & MSE~~MAE                    & MSE~~MAE          & MSE~~MAE          & MSE~~MAE    & MSE~~MAE   & MSE~~MAE      & MSE~~MAE   \\
\midrule[1.0pt]
	 {ETTh$_1$}$\rightarrow${ETTh$_2$} & 	\textbf{0.319}	~~\textbf{0.375}	 & 	\underline{0.341} ~~\underline{0.387} &  0.403 ~~0.417 	     &   0.384 ~~0.409   &   0.406~~0.422    &   0.380~~0.405      &   0.421~~0.431   &    0.380 ~~0.401 \\
\midrule
{ETTh$_1$}$\rightarrow${ETTm$_2$}   &   \textbf{0.294}	~~\underline{0.352} &   \underline{0.294}	~~ \textbf{0.350}  &   0.325~~0.360    &   0.317 ~~0.370   &   0.325 ~~0.363   &   0.314 ~~0.360       &   0.327~~0.361    &   0.316~~0.356 \\
\midrule
{ETTh$_2$}$\rightarrow${ETTh$_1$}  &   \textbf{0.487}	~~\textbf{0.482} &   \underline{0.550}~~ 0.513   &   0.669 ~~0.560   &   0.663 ~~0.540   &   0.757 ~~0.578   &   0.565 ~~0.513      &   0.865 ~~0.621   &   0.576 ~~\underline{0.511} \\
 \midrule
{ETTh$_2$}$\rightarrow${ETTm$_2$}     &   \underline{0.297}	~~\underline{0.353} &   \textbf{0.292} ~~\textbf{0.350}   &   0.327 ~~0.363   &   0.339 ~~0.371   &   0.335~~ 0.370   & 0.325 ~~0.365  &   0.342 ~~0.376   &   0.329 ~~0.371 \\
\midrule
{ETTm$_1$}$\rightarrow${ETTh$_2$}   &   \textbf{0.365}	~~\textbf{0.400} &   \underline{0.389} ~~\underline{0.417}   &   0.442 ~~0.439   &   0.440 ~~0.449   &   0.433 ~~0.439   &   0.439 ~~0.438      &   0.457~~0.454    & 0.428~~0.431 \\
\midrule
{ETTm$_1$}$\rightarrow${ETTm$_2$}   &   \textbf{0.263}	~~\textbf{0.318} &   \underline{0.268} ~~\underline{0.323}   &   0.304 ~~0.347   &   0.311 ~~0.343   &   0.313 ~~0.348   &   0.296~~0.334      &   0.322~~0.354    &   0.297~~0.330 \\
\midrule
{ETTm$_2$}$\rightarrow${ETTh$_2$}   &   \textbf{0.343}	~~\textbf{0.386} &   \underline{0.372}~~ \underline{0.408}   &   0.406 ~~0.429   &   0.429 ~~0.448   &   0.435 ~~0.443   & 0.409~~ 0.425       &   0.435 ~~0.443   &   0.417 ~~0.428 \\
\midrule
{ETTm$_2$}$\rightarrow${ETTm$_1$}  &    \textbf{0.453}	~~\textbf{0.443} &   \underline{0.496}~~ \underline{0.462}   &   0.622 ~~0.532   &   0.588 ~~0.503   &   0.769 ~~0.567   &   0.568 ~~0.492  &   0.769 ~~0.567   &   0.527 ~~0.474 \\
\bottomrule[1.0pt]
\end{tabular}
\end{table*}

\begin{table*}[t]
\centering
\setlength{\tabcolsep}{2pt}
\fontsize{9pt}{9pt}\selectfont
\centering
\caption{Short term forecasting results on M4 dataset. The weighted average values of SMAPE, MASE and OWA are reported from different sampling intervals. Lower values indicate better performance. Bold: the best, Underline: the second best.}
\label{tab:exp.shorttermforecasting}
\begin{tabular}{cccccccccc}
\toprule[1.0pt]
\multirow{3}{*}{Metric}  & \multicolumn{9}{c}{Method}     \\
\cmidrule{2-10}
         & {NeST (Ours)} & {$S^{2}$TS-LLM} & {$S^{2}$IP-LLM} & {Time-LLM} & {OFA} &  {PatchTST} & TimesNet & {DLinear} & N-HiTS \\
\midrule[1.0pt]
SMAPE		 & 	\textbf{11.940}	 & 	\underline{11.979}	 & 	12.021 	 & 	12.494 	 & 	12.690 	 & 	12.059 	 & 	12.880 &    13.639 	& 12.035 \\
\midrule
MASE	 	 & 	\textbf{1.607}	 & 	\underline{1.610}	 & 	1.612 	 & 	1.731 	 & 	1.808 	 & 	1.623 	 & 	1.836    & 2.095 & 1.625  \\
\midrule
OWA	  & 	\underline{0.860}	 & 	0.862    & \textbf{0.857}	 & 	0.913	 & 	0.940	 & 	0.869	 & 	0.955	 & 	1.051  & 0.869 	 \\
\bottomrule[1.0pt]

\end{tabular}
\end{table*}

\begin{table*}[t]
\centering
\setlength{\tabcolsep}{2pt}
\fontsize{9pt}{9pt}\selectfont
\centering
\caption{Distributed Photovoltaic (PV) power forecasting results on 5 PV stations from time series dataset 1 \cite{yao2021photovoltaic}. The look‑back window length is set to 48, with a prediction horizon of 24. RMSE and R$^2$ score are reported. Lower RMSE values and higher R$^2$ score values indicate better performance. Bold: the best, Underline: the second best.}
\label{tab:exp.solarforecasting1}
\begin{tabular}{cccccc}
\toprule[1.0pt]
Station & Station 1 & Station 2 & Station 4 & Station 7 & Station 8     \\
\midrule
Metric & RMSE~~$R^2$ & RMSE~~$R^2$ & RMSE~~$R^2$ & RMSE~~$R^2$ & RMSE~~$R^2$ \\
\midrule
NeST (Ours) & \textbf{2.651}~~\textbf{0.814}   & \textbf{2.127}~~\textbf{0.778}     & \textbf{3.905}~~\textbf{0.748}  & \textbf{2.362}~~\textbf{0.777} & \textbf{2.416}~~\textbf{0.776} \\
\midrule
RAG4PVF & \underline{2.811} ~~\underline{0.777} &  \underline{2.133}~~ \underline{0.774} &  \underline{3.987}~~ \underline{0.726} &  \underline{2.568}~~ \underline{0.725} &  \underline{2.530}~~\underline{0.712} \\
\midrule
OFA  &    2.891 ~~0.771 & 2.236 ~~0.751 & 4.101~~ 0.709 & 2.582 ~~0.716 & 2.679~~ 0.708 \\
\midrule
PatchTST  &   2.919 ~~0.758 & 2.160 ~~0.768 & 4.200 ~~0.696 & 2.598 ~~0.717 & 2.653 ~~0.711 \\
\midrule
DLinear  &  2.962~~0.749 & 2.283 ~~0.746 & 4.411 ~~0.719 & 2.661~~0.714 & 2.830 ~~0.690\\
\midrule
Informer  &   2.989~~0.750 &  2.260~~0.743   & 4.223~~ 0.692 &   2.681  ~~0.700 & 2.759 ~~ 0.691 \\
\midrule
Transformer & 2.948~~0.756    & 2.255~~0.744     & 4.447~~ 0.658      & 2.601~~ 0.718      & 2.563  ~~0.704 \\
\midrule
LSTM & 3.044~~0.755  &   2.214~~0.721  & 4.123~~0.687 & 2.665~~ 0.709 & 2.862  ~~0.688 \\
\bottomrule[1.0pt]
\end{tabular}
\end{table*}

\begin{table*}[t]
\centering
\setlength{\tabcolsep}{2pt}
\fontsize{9pt}{9pt}\selectfont
\centering
\caption{Distributed Photovoltaic (PV) power forecasting results on 4 PV stations from time series dataset 2 \cite{chen2022solar}. The look‑back window length is set to 48, with a prediction horizon of 24. RMSE and R$^2$ score are reported. Lower RMSE values and higher R$^2$ score values indicate better performance. Bold: the best, Underline: the second best.}
\label{tab:exp.solarforecasting2}
\begin{tabular}{ccccc}
\toprule[1.0pt]
Station & Station 1 (50MW) & Station 2 (130MW) & Station 5 (110MW) & Station 6 (35MW) \\
\midrule
Metric & RMSE~~$R^2$ & RMSE~~$R^2$ & RMSE~~$R^2$ & RMSE~~$R^2$ \\
\midrule
NeST (Ours) & \textbf{5.219}~~\textbf{0.853}   & \textbf{10.301}~~\textbf{0.851}     & \textbf{12.114}~~\textbf{0.723} & \textbf{3.859}~~\textbf{0.798} \\
\midrule
RAG4PVF & \underline{5.318}~~\underline{0.850} & \underline{10.313}~~\underline{0.846} & \underline{12.595}~~\underline{0.701} & \underline{3.899}~~\underline{0.792}   \\
\midrule
OFA  &   5.733~~0.825 & 11.393~~0.812 & 12.783~~0.689 & 4.350~~0.742 \\
\midrule
PatchTST  &   5.889 ~~0.816 & 10.392 ~~0.844 & 12.675 ~~0.694 & 4.033~~0.778  \\
\midrule
DLinear  &   6.218 ~~0.797 & 10.825 ~~0.833 & 13.682 ~~0.652 & 4.112~~0.771 \\
\midrule
Informer  &  5.537 ~~0.817 & 10.708 ~~0.834 & 12.989 ~~0.673 & 4.016 ~~0.768 \\
\midrule
Transformer & 5.593 ~~0.814 & 11.286 ~~0.816 & 13.754 ~~0.640 & 4.059 ~~0.775 \\
\midrule
LSTM &  5.902 ~~0.771 & 11.430 ~~0.811 & 15.281 ~~0.561 & 4.742 ~~0.693 \\
\bottomrule[1.0pt]
\end{tabular}
\end{table*}

\begin{table}[]
\caption{Ablation studies on ETTh1 dataset for [96, 192] prediction horizons. Additionally, results for $10\%$ few-shot forecasting on ETTh1 dataset for [96, 192] prediction horizons are included. MSE values are reported. Components are denoted as T: Neighborhood-aware text prototype learning, N: NNCL based objective for cross modal alignment, and P: Prototype conditioned temporal modulation.}
\fontsize{8pt}{8pt}\selectfont
\centering
\begin{tabular}{ccccccc}
\toprule[1.0pt]
T & N & P & {96}  & {192} & {10\%-96 } & {10\%-192}  \\
\midrule[1.0pt]
$\times$ & $\times$ & $\times$ &	0.411	 & 	0.427 & 0.495 & 0.572	\\
$\checkmark$ & $\times$ & $\times$ &	0.370	 & 	0.399 & 0.441 & 0.556 	\\
$\times$ & $\checkmark$ & $\times$ &	0.364	 & 	0.397 & 0.437 & 0.545  \\\
$\checkmark$ & $\checkmark$ & $\times$ &	0.364	 & 	0.395 & 0.428 & 0.510  	 \\
\midrule
$\checkmark$ & $\checkmark$ & $\checkmark$ & \textbf{0.356} 	 & 	\textbf{0.391} 	 & 	\textbf{0.414}   &   \textbf{0.482} \\
\bottomrule[1.0pt]
\end{tabular}
\label{tab:exp.ablations}
\end{table}

\subsection{Evaluation Results}
\subsubsection{Long-term Forecasting Results}
Long-term forecasting results are presented in Table \ref{tab:exp.longtermforecasting}. In terms of MSE, our method outperforms the second best methods by approximately 2.4\% on ECL (0.160 vs. 0.164), 1.0\% on Traffic (0.392 vs. 0.396), 2.2\% on ETTh$_1$ (0.395 vs. 0.404), 0.6\% on ETTh$_2$ (0.319 vs. 0.321), and 2.7\% on ETTm$_2$ (0.251 vs. 0.258). Averaged across datasets, this corresponds to an overall relative reduction of 1.2\% in MSE compared to second best existing method. A similar trend is observed for MAE, where NeST consistently attains lower errors, yielding an average reduction of 1.3\% over competing approaches. These results indicate that the proposed method provides consistent improvements across diverse long-term time series forecasting settings, outperforming both classical and LLM-based baselines.

\subsubsection{Few-shot Forecasting Results}
Few-shot forecasting results are shown in Table \ref{tab:exp.fewshotforecasting}, where only 10\% of the training data is used for finetuning. As demonstrated in results, NeST achieves the best results on three out of four datasets, demonstrating its effectiveness under data scarce conditions. Specifically, it improves MSE by 1.3\% on ETTh$_1$ (0.536 vs. 0.543), 2.4\% on ETTh$_2$ (0.363 vs. 0.372), and 0.7\% on ETTm$_2$ (0.277 vs. 0.279) compared to the second best methods. On ETTm$_1$, NeST remains highly competitive, achieving 0.417 MSE, closely matching the best performing baseline (0.414). Similar trends are observed in MAE, where NeST either achieves the best or second best results across all datasets. Overall, the proposed method demonstrates consistent performance improvements, indicating that it can effectively leverage limited supervision available in data scarce scenarios and maintain robust forecasting performance.

\subsubsection{Zero-shot Forecasting Results}
Table \ref{tab:exp.zeroshotforecasting} summarizes the zero‑shot forecasting performance under cross‑dataset transfer settings. NeST outperforms existing baseline methods in seven out of eight transfer scenarios, demonstrating strong cross‑domain generalization without task specific finetuning. When trained on ETTh$_1$ and evaluated on ETTh$_2$, NeST achieves a 6.5\% reduction in MSE relative to the second best method, while the reverse transfer (ETTh$_2$ $\rightarrow$ ETTh$_1$) yields an reduction of 11.5\%. Similarly, training on ETTm$_1$ and testing on ETTh$_2$ and ETTm$_2$ results in MSE reductions of 6.2\% and 1.9\%, respectively. Strong performance is also observed when the model is trained on ETTm$_2$, where NeST surpasses competing methods by 7.8\% on ETTh$_2$ and 8.7\% on ETTm$_1$. These results indicate that NeST effectively transfers learned temporal and semantic representations across datasets with differing sampling frequencies and statistical characteristics. Such robustness is particularly valuable in real‑world forecasting scenarios, where labeled data may be limited or unavailable, and models must generalize reliably to previously unseen domains.

\subsubsection{Short-term Forecasting Results}
As presented in Table \ref{tab:exp.shorttermforecasting}, NeST demonstrates competitive performance across all evaluation metrics on the M4 short-term forecasting benchmark. It achieves the best SMAPE of 11.940, improving over the second best result (11.979) by 0.3\%, and also attains the lowest MASE of 1.607, outperforming the next best (1.610) by around 0.2\%. In terms of OWA, NeST achieves 0.860, which is close to the existing best performing method (0.857), remaining within a 0.3\% margin. Overall, these results indicate that NeST maintains strong and stable performance in short-term forecasting, matching or surpassing SOTA methods across multiple metrics while demonstrating robustness across different sampling intervals.

\subsubsection{Distributed Photovoltaic (PV) Power Forecasting Results}
We further evaluate NeST on two real‑world distributed PV power forecasting datasets collected from independent sources, covering a total of nine PV stations. As reported in Tables \ref{tab:exp.solarforecasting1} and \ref{tab:exp.solarforecasting2}, NeST consistently outperforms existing SOTA methods across both datasets. For the first dataset (Table \ref{tab:exp.solarforecasting1}), which includes five PV stations, our method achieves the highest average performance, surpassing the second best method; an LLM based approach specifically designed for PV power forecasting by 4.9\% on average in terms of R$^2$ score. Similarly, results on the second dataset (Table \ref{tab:exp.solarforecasting2}), comprising four PV stations, demonstrate that NeST attains superior performance, outperforming the second best baseline by an average of 1.2\% in R$^2$ score. These results highlight the practical effectiveness of NeST in distributed PV power forecasting, a challenging real‑world application characterized by strong variability and limited labeled data. Notably, unlike competing methods tailored explicitly for solar forecasting, NeST achieves these gains without any architectural modifications, while maintaining strong performance across diverse time series benchmark datasets. This demonstrates the generalizability and robustness of the proposed method, making it suited for deployment in real‑world renewable energy forecasting scenarios.

\begin{figure*}
  \centering
  \includegraphics[width=\linewidth]{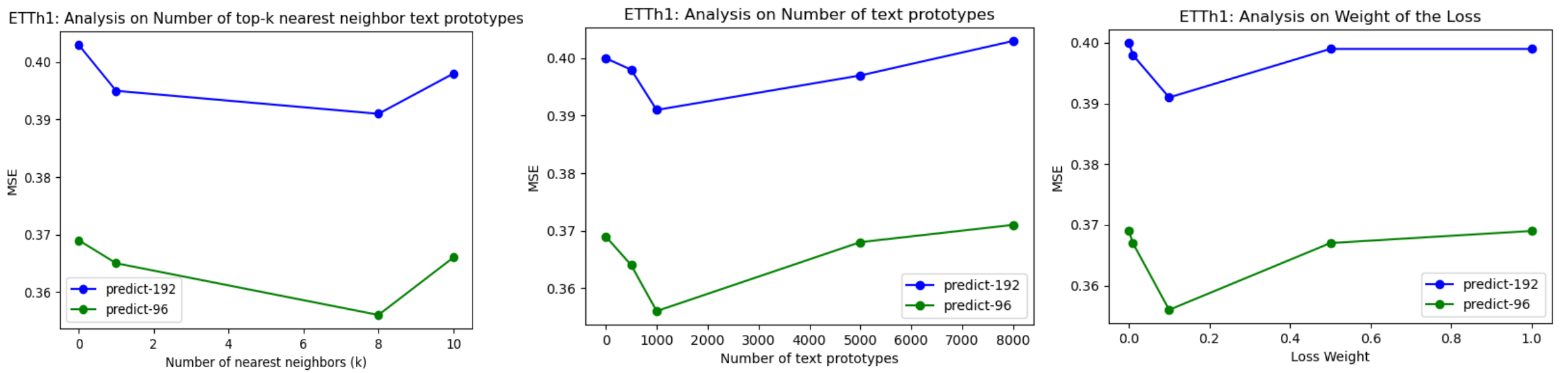}
  \caption{Illustration of the analysis of parameter sensitivity. MSE values are reported for ETTh1 dataset for [96, 192] prediction
horizons. Analysis on (a) Number of top-k nearest text prototypes, (b) Number of text prototypes, and (c) Weight of the loss are demonstrated.}
  \label{fig:sensitivityanalysis}
\end{figure*}

\begin{figure*}
  \centering
  \includegraphics[width=\linewidth]{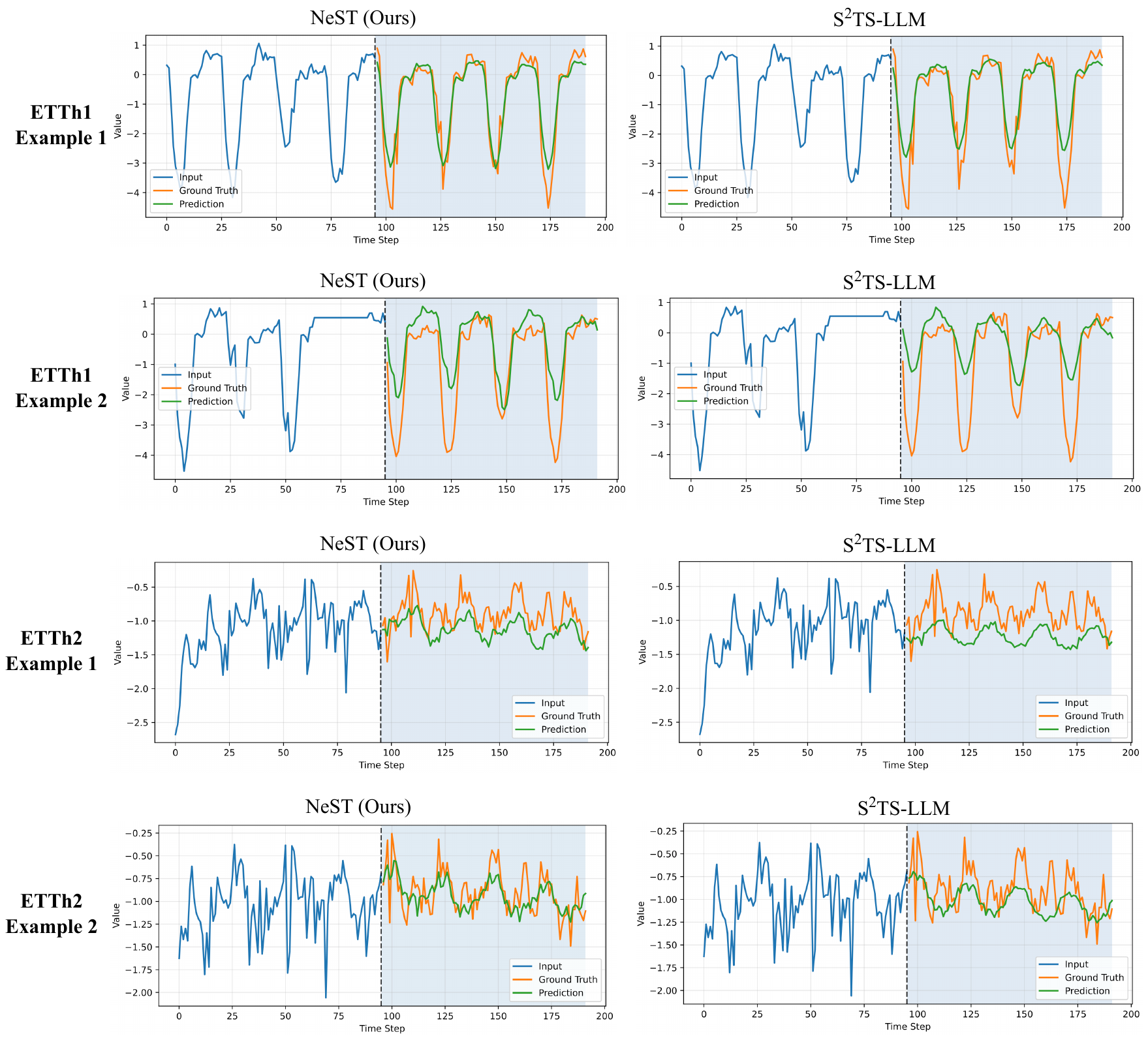}
  \caption{Comparative prediction plots visualization of long term forecasting results on the 96 prediction horizon for ETTh$_1$ and ETTh$_2$ datasets with 96 look-back length. Ground truth of look back length (blue), predictions (green), and ground truth of prediction horizon (orange) are shown for our model alongside results from {$S^{2}$TS-LLM \cite{s2tsllm}}.}
  \label{fig:predictionplots}
\end{figure*}

\begin{table*}[t]
\centering
\setlength{\tabcolsep}{2pt}
\fontsize{8pt}{8pt}\selectfont
\centering
\caption{Performance comparison of different methods across varying look-back lengths. 96 is used as the prediction horizon.}
\label{tab:exp.lookbacklengths}
\begin{tabular}{ccccccc}
\toprule[1.0pt]
\multirow{3}{*}{Dataset}  & Look-back Length & 512 & 336 & 192 & 96 & 48     \\
\cmidrule{2-7}
& Metric & MSE~~MAE & MSE~~MAE & MSE~~MAE & MSE~~MAE & MSE~~MAE \\
\midrule
\multirow{5}{*}{ETTh$_1$} & NeST (Ours) & \textbf{0.356}~~\textbf{0.390} & \textbf{0.362}~~\textbf{0.388} & \textbf{0.371}~~\textbf{0.389 } & \textbf{0.374}~~\textbf{0.391} & \textbf{0.375}~~\textbf{0.394} \\
\cmidrule{2-7}
& {$S^{2}$TS-LLM \cite{s2tsllm}} & 0.355~~0.387 & 0.365~~0.389 & 0.372~~0.391 & 0.378~~0.391 & 0.398~~0.411 \\
\cmidrule{2-7}
& {$S^{2}$IP-LLM \cite{pan2024s}} & 0.368~~0.403 & 0.367~~0.398 & 0.368~~0.393 & 0.379~~0.394 & 0.390~~0.401 \\
\cmidrule{2-7}
& {OFA \cite{zhou2023one}} & 0.376~~0.397 & 0.378~~0.399 & 0.380~~0.401 & 0.383~~0.401 & 0.385~~0.403 \\
\midrule
\multirow{5}{*}{ETTh$_2$} & NeST (Ours) & \textbf{0.260}~~\textbf{0.326} & \textbf{0.272}~~\textbf{0.333} & \textbf{0.279}~~\textbf{0.335} & \textbf{0.279}~~\textbf{0.333} & \textbf{0.289}~~\textbf{0.337} \\
\cmidrule{2-7}
& {$S^{2}$TS-LLM \cite{s2tsllm}} & 0.260~~0.329 & 0.281~~0.335 &  0.288~~0.338 & 0.291~~0.336 & 0.310~~0.350 \\
\cmidrule{2-7}
& {$S^{2}$IP-LLM \cite{pan2024s}} & 0.284~~0.345   &  0.282~~0.343   &  0.288~~0.343   &  0.293~~0.343   &  0.301~~0.344 \\
\cmidrule{2-7}
& {OFA \cite{zhou2023one}}  &    0.285~~0.342 & 0.286~~0.345 & 0.294~~0.349 & 0.295~~0.349 & 0.299~~0.350 \\
\bottomrule[1.0pt]
\end{tabular}
\end{table*}

\subsubsection{Ablation Studies}
We conduct ablation studies to analyze the contribution of each core component of NeST. Experiments are performed on the ETTh$_1$ dataset with prediction horizons of 96 and 192, following the long-term forecasting setup. In addition, we report results under 10\% few-shot training setting on ETTh$_1$ dataset to examines the effectiveness of individual components in data scare conditions. NeST comprises of three key components: neighborhood‑aware text prototype learning (T), nearest‑neighbor contrastive learning (N), in which a support set is constructed using the learned text prototypes, and text prototype conditioned temporal modulation (P), which integrates semantic information into time series patch embeddings without relying on direct token concatenation. In the ablation experiments, we progressively incorporate these components, evaluating models with only one component enabled and comparing them against the baseline model without any of the proposed components. The results in Table \ref{tab:exp.ablations} demonstrate that each component contributes positively to forecasting performance, with particularly pronounced gains in the few‑shot setting. For the 192‑step forecast horizon with only 10\% of training data, introducing neighborhood-aware text prototype learning reduces MSE by 2.8\%, while the addition of the NNCL component yields a 4.7\% reduction in MSE compared to the baseline model. Furthermore, incorporating text prototype conditioned temporal modulation reduces MSE by an additional 5.5\% compared to the variant with the neighborhood-aware text prototype learning and NNCL based alignment. These findings indicate that the synergistic combination of all proposed components is critical for effectively adapting LLMs to time series forecasting, especially in data scarce settings, where NeST exhibits the greatest relative improvements.

\subsubsection{Parameter Sensitivity Analysis}
To investigate the robustness of the proposed method, we conducted a sensitivity analysis on three key hyperparameters as presented: (a) the number of nearest-neighbor text prototypes; $k$, (b) the number of learnable text prototypes; $U$, and (c) the weight associated with the custom loss component; $\lambda$. Experiments were performed on each dataset and sensitivity analysis on ETTh$_1$ dataset using prediction horizons of 96 and 192, with MSE as the evaluation metric is presented in Fig. \ref{fig:sensitivityanalysis} as a representative example. We observe that increasing the number of nearest-neighbor text prototypes ($k$) initially leads to improved forecasting performance for both prediction horizons, with the lowest MSE obtained around $k=8$. This behavior suggests that incorporating a moderate number of neighboring prototypes provides richer semantic context during nearest-neighbor contrastive learning. However, performance slightly deteriorates when $k$ is increased further, indicating that distant prototypes introduce less relevant semantic information and may weaken the discriminative capability of the alignment process. Fig. \ref{fig:sensitivityanalysis}(b) illustrates the effect of varying the number of learnable text prototypes ($U$). The results show that a relatively compact prototype set yields the best forecasting accuracy. Increasing the number of prototypes beyond this point results in a gradual increase in forecasting error. This suggests that a smaller set of prototypes is sufficient to capture the dominant semantic regions in the pretrained word token embedding space. In contrast, a larger prototype set increases model complexity and may introduce redundant or highly similar prototype representations, reducing the performance. Fig. \ref{fig:sensitivityanalysis}(c) investigates the sensitivity of the framework to the loss weight ($\lambda$). When $\lambda=0$, the neighborhood-aware prototype learning and NNCL objectives are removed, leading to inferior forecasting performance. Introducing a small positive weight significantly improves accuracy, demonstrating the effectiveness of enforcing semantic structure preservation and cross-modal alignment. The best performance is achieved at $\lambda=0.1$. As $\lambda$ increases further, forecasting performance gradually degrades because the optimization process becomes overly constrained by the auxiliary objectives, reducing emphasis on the primary forecasting task. This indicates that an appropriate balance between forecasting supervision and representation regularization is essential for optimal performance. Overall, the proposed NeST framework demonstrates stable behavior across a broad range of hyperparameter values, indicating strong robustness to parameter selection.

\subsubsection{Effect of Look-back Window Length}
Table \ref{tab:exp.lookbacklengths} analyses the performance of NeST under varying look‑back window lengths with a fixed prediction horizon of 96. Across both ETTh$_1$ and ETTh$_2$ datasets, NeST consistently demonstrates competitive or superior performance compared with existing LLM-based baselines, highlighting its robustness to changes in historical context length. For ETTh$_1$, NeST achieves the lowest MSE and MAE for look‑back windows of 336, 192, 96, and 48, outperforming the existing competing method by up to 2.6\% in MSE and 1.7\% in MAE, particularly in shorter context settings. While performance is comparable for the longest look‑back window (512), NeST remains highly competitive, matching or closely approaching the current baseline. A similar trend is observed on ETTh$_2$, where NeST achieves consistent improvements across all look‑back lengths, with up to 4.1\% relative MSE reduction compared with the second‑best method. Notably, the performance gains are more pronounced as the look‑back window shortens, indicating that NeST effectively leverages semantically informed text prototypes and cross-modal alignment to compensate for limited historical information. These results demonstrate that NeST is robust to variations in input context length and maintains strong forecasting accuracy even under constrained temporal histories, which is particularly important for practical time series forecasting scenarios where long historical windows may not be available.

\subsubsection{Prediction Visualizations}
To provide qualitative insights into the forecasting behavior of NeST, Figure \ref{fig:predictionplots} presents representative prediction examples from the ETTh$_1$ and ETTh$_2$ datasets, comparing our method with the second best existing baseline, S$^2$TS-LLM \cite{s2tsllm}. Four forecasting instances are selected to illustrate the ability of both models to capture temporal dynamics under different patterns and distinct levels of variability. For the ETTh$_1$ dataset, even though both methods successfully capture the overall periodic behavior of the series, our method demonstrates a closer alignment with the ground truth trajectories, particularly around turning points and peak values. In Example 1, NeST more accurately follows the amplitude variations across multiple cycles, whereas S$^2$TS-LLM exhibits larger deviations around local maxima and minima. Similar behavior can be observed in Example 2, where NeST produces forecasts that more closely match the timing and magnitude of the observed fluctuations. In contrast, S$^2$TS-LLM tends to smooth the signal and underestimates several extreme values, resulting in larger prediction errors. The qualitative differences become more evident on the ETTh$_2$ dataset, which contains noisier and less regular temporal patterns. In both examples, NeST is able to better capture the short term oscillations and local variations present in the ground truth series. Although both methods exhibit some degree of smoothing, NeST consistently generates predictions that remain closer to the actual observations throughout the forecasting horizon. In contrast, S$^2$TS-LLM produces flatter forecasts with reduced responsiveness to local changes, leading to noticeable underestimation of peaks and over-smoothing of temporal dynamics. Overall, the visualizations demonstrate that while S$^2$TS-LLM remains a strong competitor, NeST provides more accurate and temporally consistent forecasts across both datasets. The proposed method better preserves the underlying temporal structure, captures amplitude variations more accurately, and adapts more effectively to both periodic and irregular patterns. These qualitative observations are consistent with the quantitative improvements, further validating the effectiveness of the proposed framework for time series forecasting.

\begin{figure}
  \centering
  \includegraphics[width=\linewidth]{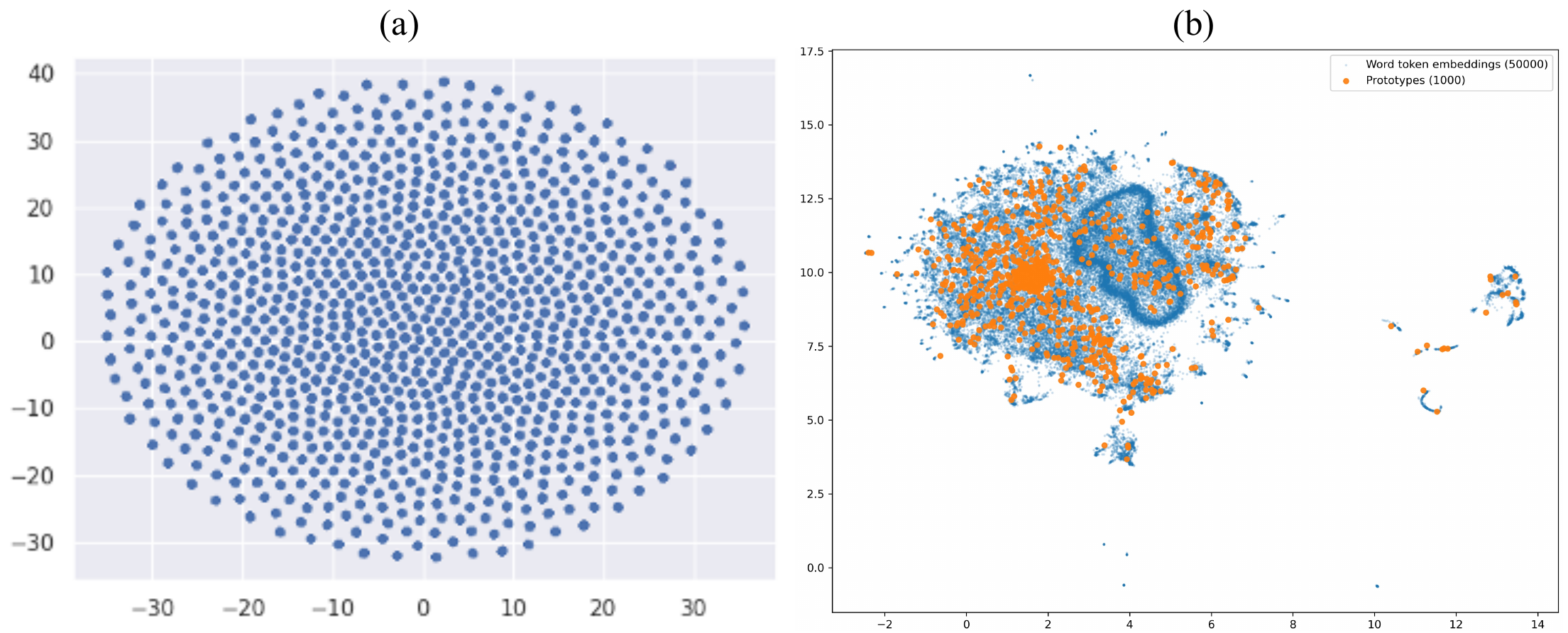}
  \caption{Illustration of UMAP visualizations of optimized text prototypes in 2-dimensional space. (a) Text prototypes obtained from existing linear probing mechanism, (b) Text prototypes obtained from proposed neighborhood-aware text prototype learning (Orange color) overlaid with pretrained word token embeddings (Blue color).}
  \label{fig:textprototypesvis}
\end{figure}

\subsubsection{Text Prototype Visualizations}
To examine the effect of the proposed neighborhood-aware text prototype learning, we obtain visualizations of the learned text prototypes. Fig. \ref{fig:textprototypesvis} presents a comparison between the learned text prototypes obtained from the existing linear probing mechanism and the text prototypes obtained from neighborhood-aware approach overlaid with the pretrained LLM word token embeddings. As shown in Fig. \ref{fig:textprototypesvis}(a), prototypes learned via linear probing form an isolated and highly regular structure that is largely detached from the pretrained word token embedding manifold. This suggests that the optimization objective primarily focus on forecasting without preserving semantic relationships acquired during LLM pretraining. In contrast, our neighborhood-aware text prototype learning approach produces prototypes that are embedded within high density regions of the word token space (Fig. \ref{fig:textprototypesvis}(b)), closely following the underlying semantic structure. Moreover, these prototypes provide broader coverage of the embedding space by spanning multiple semantic clusters, resulting in more diverse and representative semantic anchors. Such representation facilitates more effective prototype retrieval during nearest-neighbor contrastive learning and strengthens the prompt formulation. Overall, the visualization provides qualitative evidence that neighborhood-aware text prototype learning better preserves pretrained semantic knowledge of the LLM than conventional linear probing, contributing to improved forecasting performance as demonstrated in the experimental evaluation results.

\section{Conclusion}
\label{sec:conclusion}
In this paper, we present NeST, a new LLM based framework for time series forecasting that addresses key limitations of existing approaches in cross-modal alignment and prompt formulation. To preserve the semantic geometry of pretrained word token embeddings, we introduced a neighborhood-aware text prototype learning strategy that learns representative semantic anchors within the LLM embedding space. To strengthen the alignment between temporal and textual representations, we proposed a nearest-neighbor contrastive learning objective. Furthermore, we developed a parameter efficient text prototype conditioned temporal modulation mechanism that adaptively injects semantic information into temporal features, avoiding the limitations of simple concatenation and the computationally heavy cross-attention based prompting. Comprehensive experiments across long-term, short-term, few-shot, and zero-shot forecasting benchmarks demonstrated the robustness of NeST over SOTA methods. Beyond standard benchmarks, the proposed framework also showed strong practical value on real-world distributed photovoltaic power forecasting datasets while consistently improving predictive performance across multiple operational sites and conditions. Future research will investigate extending the framework to better capture inter-channel dependencies in multivariate time series and further enhance temporal representation learning.

\section{Declaration of generative AI and AI-assisted technologies in the manuscript preparation process}
During the preparation of this work the authors used ChatGPT in order to improve language and readability. After using this tool/service, the authors reviewed and edited the content as needed and take full responsibility for the content of the article.

\bibliographystyle{cas-model2-names}

\bibliography{cas-refs}

\end{document}